\documentclass[letterpaper, 10 pt, conference]{ieeeconf}  %

\IEEEoverridecommandlockouts               

\def\BibTeX{{\rm B\kern-.05em{\sc i\kern-.025em b}\kern-.08em
    T\kern-.1667em\lower.7ex\hbox{E}\kern-.125emX}}

\usepackage[english]{babel}
\usepackage[utf8]{inputenc}
\usepackage{graphicx}
\usepackage{subfig} 
\usepackage{amsmath}
\usepackage{amsthm}
\usepackage{amsfonts}
\usepackage{amssymb}
\usepackage{amsbsy}
\usepackage{siunitx}

\usepackage{algorithm}
\usepackage{algorithmic}
\usepackage{yaacro}
\usepackage{bm}

\usepackage{xcolor}
\usepackage{graphicx}
\usepackage{subfig}
\usepackage{wrapfig}
\usepackage{booktabs} %
\usepackage{subfiles}

\usepackage{multicol}
\usepackage{latexsym} 
\usepackage[hyphens]{url}
\let\labelindent\relax
\usepackage{enumitem}
\usepackage{array}
\usepackage{bbm}

\usepackage{xspace}
\usepackage{xcolor}

\usepackage[normalem]{ulem}

\newcommand{\eg}{\textit{e.g.}}

\begin{acgroupdef}[list=acronimos]

\end{acgroupdef}

\definecolor{wine}{RGB}{204, 0, 102}
\definecolor{ocean}{RGB}{13, 121, 202}
\definecolor{light_ocean}{RGB}{18, 178, 235}
\definecolor{dark_ocean}{RGB}{10, 89, 148}
\definecolor{grey}{RGB}{170, 170, 170}
\definecolor{light-grey}{RGB}{220, 220, 220}
\definecolor{dark_gray}{rgb}{0.2, 0.2, 0.2} 
\definecolor{med-grey}{rgb}{0.3, 0.3, 0.3} 
\definecolor{grape}{RGB}{112,48,160}
\definecolor{aqua}{RGB}{52,172,139}
\definecolor{dark_aqua}{RGB}{35,115,93}
\definecolor{dark_orange}{RGB}{216,92,0}
\definecolor{vibrant_orange}{RGB}{255, 102, 0}
\definecolor{vibrant_blue}{RGB}{14, 120, 255}
\definecolor{vibrant_pink}{RGB}{255, 0, 104}
\definecolor{dark_red}{RGB}{122, 0, 0}
\definecolor{dark_green}{RGB}{0, 92, 34}

\usepackage[
  bookmarks=false,
  pdfpagelabels=false,
  hyperfootnotes=false,
  hyperindex=false,
  colorlinks,urlcolor= orange,
  citecolor=orange,
  linkcolor=orange,
]{hyperref}

\usepackage{tikz}
\usetikzlibrary{shapes}
\usetikzlibrary{calc} 
\usepackage[compact]{titlesec}
\titlespacing{\paragraph}{%
  0pt}{%
  0.2\baselineskip}{%
  .5em}%
\usepackage{pgfplots}
\usepgfplotslibrary{fillbetween}	

\usepackage{xcolor,pifont}
\newcommand*\colourcheck[1]{%
  \expandafter\newcommand\csname #1check\endcsname{\textcolor{#1}{\ding{52}}}%
}
\newcommand{\xmark}{\ding{55}}%
\colourcheck{blue}
\colourcheck{green}
\colourcheck{red}
\colourcheck{black}

\newcommand{\para}[1]{\medskip\noindent\textbf{#1. }}

\newcommand{\safelang}{\textcolor{orange}{\textbf{Safe-Lang}}\xspace}
\newcommand{\mppilang}{\textcolor{grape}{\textbf{Plan-Lang}}\xspace}
\newcommand{\safe}{\textcolor{dark_aqua}{\textbf{Safe-SLAM}}\xspace}
\newcommand{\mppi}{\textcolor{black}{\textbf{Plan-SLAM}}\xspace}

\newcommand{\state}{s}

\newcommand{\robot}{\mathcal{R}}

\newcommand{\env}{E}

\newcommand{\dyn}{f}

\newcommand{\dynR}{\dyn}

\newcommand{\dsR}{\dot{\state}}

\newcommand{\aR}{a}

\newcommand{\dstbR}{d}
\newcommand{\langCmd}{\ell}
\newcommand{\langSpace}{\mathcal{L}}

\newcommand{\ctrlSetR}{\mathcal{A}}
\newcommand{\dstbSetR}{\mathcal{D}}
\newcommand{\stateSpace}{\mathcal{S}}

\newcommand{\obs}{o}
\newcommand{\obsSpace}{\mathcal{O}}

\newcommand{\pred}{\mathcal{P}}
\newcommand{\vlm}{\phi}
\newcommand{\distThresh}{\tau_{dist}}
\newcommand{\bbox}{\mathcal{B}}
\newcommand{\image}{\mathcal{I}}

\newcommand{\policyR}{\pi_\robot}

\newcommand{\overallpolicyR}{\pi_\robot^*}

\newcommand{\failureset}{\mathcal{F}}

\newcommand{\marginfunc}{g}

\newcommand{\valfunc}{V}
\newcommand{\safeSet}{\mathcal{S}^\shield}

\newcommand{\tdummy}{\tau}

\usepackage{fontawesome5}
\newcommand{\policy}{\pi}
\newcommand{\shield}{\text{\tiny{\faShield*}}}

\newcommand{\fallback}{\policy^{\shield}_\robot}

\begin{document}
\title{\LARGE \bf Updating Robot Safety Representations\\ Online from Natural Language Feedback\\
\thanks{$^*$ Equal Contribution. $^\dagger$ Equal Advising.}
\thanks{\textsuperscript{1}School of Engineering, Federal University of Minas Gerais, Brazil. Email: \texttt{leohmcs@ufmg.br}. Work done during Robotics Institute Summer Scholars (RISS) Program at Carnegie Mellon University. \textsuperscript{2}Electrical and Computer Engineering, University of Rochester. Email: \texttt{zli133@u.rochester.edu}. 
\textsuperscript{3}Department of Cognitive Robotics, Delft University of Technology. Email: \texttt{l.peters@tudelft.nl}.
\textsuperscript{4}Electrical and Computer Engineering, University of Southern California. Email: \texttt{somilban@usc.edu}. \textsuperscript{5}Robotics Institute, Carnegie Mellon University. Email: \texttt{{abajcsy}@cmu.edu}}
}

\author{Leonardo Santos\textsuperscript{1*},\quad Zirui Li\textsuperscript{2*},\quad Lasse Peters\textsuperscript{3},\quad Somil Bansal\textsuperscript{4$\dagger$},\quad Andrea Bajcsy\textsuperscript{5$\dagger$}
}

\makeatletter
\let\@oldmaketitle\@maketitle%
\renewcommand{\@maketitle}{\@oldmaketitle%
\setcounter{figure}{0} %
\centering
\includegraphics[width=1.0\textwidth]{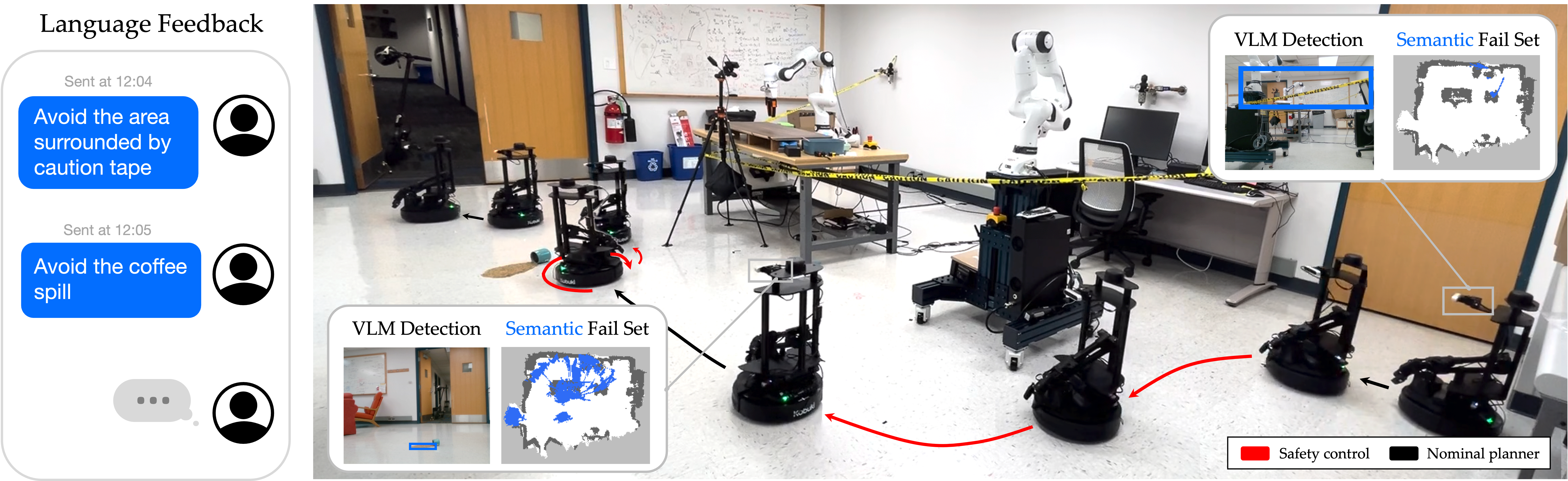}
\captionof{figure}{Natural language provides an intuitive interface for people to specify constraints they care about online, like restricted areas behind caution tape or coffee spills. We leverage advances in vision-language models to interpret multimodal language and image data, infer semantically-meaningful constraints, and update robot safety controllers online. Video results and code at the project website: \href{https://cmu-intentlab.github.io/language-informed-safe-navigation/}{https://cmu-intentlab.github.io/language-informed-safe-navigation/}.}
\label{fig:front-fig}
\vspace{-0.3in}
\bigskip}
\makeatother
\maketitle
\pdfminorversion=4

\begin{abstract}
Robots must operate safely when deployed in novel and human-centered environments, like homes. 
Current safe control approaches typically assume that the safety constraints are known \textit{a priori}, and thus, the robot can pre-compute a corresponding safety controller. 
While this may make sense for some safety constraints (\eg, avoiding collision with walls by analyzing a floor plan), other constraints are more complex (\eg, spills), inherently personal, context-dependent, and can only be identified at deployment time when the robot is interacting in a specific environment and with a specific person (\eg, fragile objects, expensive rugs).
Here, language provides a flexible mechanism to communicate these evolving safety constraints to the robot.
In this work, we use vision language models (VLMs) to interpret language feedback and the robot's image observations to continuously update the robot's representation of safety constraints. 
With these inferred constraints, we update a Hamilton-Jacobi reachability safety controller online via efficient warm-starting techniques. 
Through simulation and hardware experiments, we demonstrate the robot's ability to infer and respect language-based safety constraints with the proposed approach. 
\end{abstract}

\section{Introduction}
As robots are increasingly integrated into human environments, ensuring their safe operation is critical.
Designing safe controllers for robots is a well-studied problem in robotics; however, the current approaches often assume that the safety constraints are known in advance, and thus, a safety controller can be synthesized offline. 
While this approach may be effective for static and well-defined constraints (e.g., walls or fixed obstacles), it is insufficient in complex, human-centered environments, where safety requirements are often personalized and context-dependent. 
For example, one may not want a cleaning robot to drive  
through a workout area during exercise, and a warehouse robot should avoid entering areas temporarily blocked with caution tape (Figure~\ref{fig:front-fig}). 

In such cases, language provides a flexible communication channel between the robot and the operator who can easily describe constraints they care about (\eg, ``Avoid the area surrounded by caution tape''). 
In this work, we develop a framework for updating robot safety representations \textit{online} through such natural language feedback. 
Our key idea is that pre-trained open-vocabulary vision-language models (VLMs) are not only a useful interface for constraint communication, but they provide an easy way to convert multimodal data observed online (RGB-D and language) into updated safety representations. 
With this, the robot can detect hard-to-encode constraints such as a workout zone, coffee spills, or designated no-go zones (Figure~\ref{fig:front-fig}). 
To ensure safety with respect to both pre-defined and new language constraints, we leverage Hamilton-Jacobi reachability analysis \cite{bansal2017hamilton, mitchell2005time} to compute a \textit{policy-agnostic} safety controller for the robot which is constantly updated online via efficient warm-starting techniques \cite{bajcsy2019efficient, herbert2019reachability}.
The safety controller intervenes only when the robot's  nominal planner is at risk of violating either the physical or semantic safety constraints, and provides a corrective safe action.
Through simulation studies and experiments on a hardware testbed, we demonstrate the ability of our framework to enable the robot to operate safely, even when new language-based constraints are introduced during deployment.

\section{Related Work} \label{sec:related-work}
\para{Language-Informed Robot Planning} 
While this topic has been explored for over a decade (see review \cite{tellex2020robots}), advances in internet-scale language and vision-language models have significantly grown language-informed robot planning approaches. 
Recent works use language for high-level semantic or motion planning \cite{ahn2022can, shah2023navigation, singh2023progprompt, liu2022structformer}, providing corrective feedback \cite{sharma2022correcting, cui2023no}, for low-level control primitives \cite{liang2023code}, and for language-conditioned end-to-end policies \cite{lynch2020language, kim2024openvla}.  
One common theme in these lines of work is that language provides a flexible mechanism to interact with the robot. Building upon this observation, we use language feedback in our work to enhance robot safety during deployment time. 

\para{Safety Constraint Inference from Human Feedback} There is a relatively smaller body of work focused on constraint learning from human feedback. Prior works have inferred state constraints offline from human demonstrations \cite{scobee2019maximum, chou2020learning, kim2024learning, lindner2024learning, shah2023learning}, and inferred constraints represented as logical (LTL or STL) specifications from natural language \cite{finucane2010ltlmop, pan2023data}. 
Our framework focuses on inferring novel state constraints \textit{online} based on multimodal data of image observations and natural language feedback. 

\para{Safety Filtering} 
Safety filters are a popular mechanism to ensure safety for autonomous robots under \textit{any} off-the-shelf planner \cite{hewing2020learning, hsu2023safety}. 
The key idea is to use a nominal planner whenever it is safe for the system and intervene with a safety-preserving action whenever the system's safety is at risk.
The most popular paradigms to construct safety filters are control barrier functions (CBFs) \cite{ames2019control, qin2021learning, chen2020guaranteed, li2023robust, dawson2022safe, liu2023safe}, Hamilton-Jacobi (HJ) reachability analysis \cite{herbert2017fastrack, singh2017robust, tian2022safety, nguyen2024gameplay}, and model predictive shielding \cite{brunke2022safe}. 
We leverage HJ reachability, as it can be easily applied to general nonlinear systems, accounts for control constraints and system dynamics uncertainty, and is associated with a suite of numerical tools \cite{mitchell2004toolbox}.
We build on prior work \cite{herbert2019reachability, bajcsy2019efficient, borquez2022parameter} which proposed algorithms for efficiently updating reachability-based safety filters online as the safety constraints change. 
Our key innovation is incorporating multimodal data of language and images to this online update. 

\section{Problem Formulation}
\label{sec:problem-statement}
\para{Robot and Environment} We model the robot as a continuous-time dynamical system $\dsR(t) = \dynR(\state, \aR, \dstbR)$, where $t \in \mathbb{R}$ is the time, $\state \in \stateSpace$ is the robot state (e.g., planar position and heading), $\aR \in \ctrlSetR$ is the robot's control input (e.g., linear and angular velocity). Here, $\dstbR \in \dstbSetR$ is the disturbance which can be an exogenous input (e.g., wind for an aerial vehicle) or represent model uncertainty (e.g., unmodelled tire friction) that we want to be robust to. 
We assume that the flow field $\dynR: \stateSpace\times \ctrlSetR \times \dstbSetR \rightarrow \stateSpace$ is uniformly continuous in time, and Lipschitz continuous in $\state$ for fixed $\aR$ and $\dstbR$. 
The robot is operating in an environment $E$ that it shares with a human, and we assume that the two agents do not expect to physically interact. We use the term ``environment'' here broadly to refer to factors that are external to the robot (e.g., a building that the robot is navigating in or the surrounding lighting conditions).
We also assume that we are given a nominal robot policy $\policyR(\state; \env)$ that maps the robot state to control inputs.  
$\policyR$ is typically designed to obtain a desired robot behavior, such as reaching a particular goal location for a navigation robot.

\para{Robot Sensor and Perception} The robot has a sensor $\sigma : \stateSpace~\times~\env~\rightarrow~\obsSpace$ that yeilds (high-dimensional) RGB-D observations. At any time $t \in [0,T]$ during the deployment horizon, let $\obs^t \in \obsSpace$ be the robot's observation. 

\para{Human Language Feedback} A human can augment the robot's constraint set at any time during deployment via language commands.
More formally, let the human's language command be denoted by $\langCmd^t \in \langSpace$, where $t \geq 0$ is any time during deployment. 
In this work, $\langSpace$ are open-vocabulary commands and the set also includes null in which case the person does not describe a new constraint. 

\para{Safety Representation: Failure Set} Let $\failureset^*_\env \subset \stateSpace$ be the failure set in the human's mind consisting of both physical constraints that are known a priori (e.g., floorplan geometry), as well as the semantically-meaningful constraints that the human describes in the language (e.g., caution tape, spill, etc.).
Intuitively, the failure set captures the state constraints that our system must avoid.
Traditionally, this failure set is assumed to be specified a priori, and then utilized to compute a safe set and corresponding safety controller automatically. 
Our work precisely aims to relax this assumption.
Thus, $\failureset^*_\env$ can change online as the robot is operating in the environment.

\para{Objective} 
We seek to design a robot controller $\overallpolicyR$ for the robot that 
respects the safety constraints $\failureset^*_\env$ at all times while following the nominal policy $\policyR$ as closely as possible.

\begin{figure*}[t!]
    \centering
    \includegraphics[width=1\linewidth]{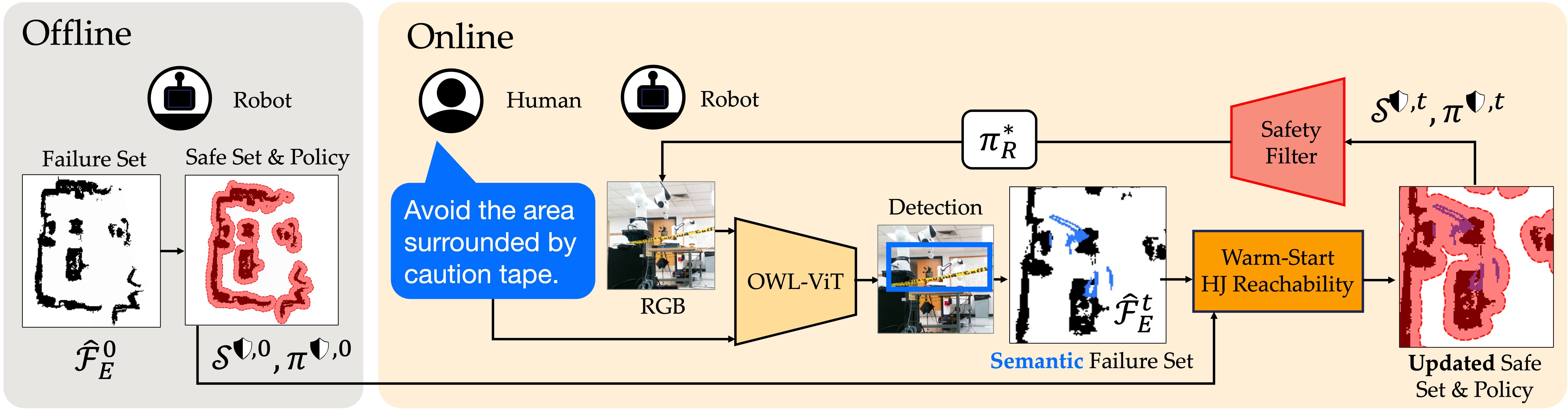}
    \caption{\textbf{Updating Robot Safety Representations Online from Language Feedback.} (left) Offline, the robot has an initial failure set ($\hat\failureset^{,0}_\env$) and computes the corresponding safe set (${\safeSet}^{,0}$) and safety policy (${\fallback}^{,0}$). (right) Online, the person describes their semantic constraint. Using a vision-language model, the robot converts the language-image data into a new failure set. This, along with the previously-computed safe set, are used to efficiently update the safety filter that shields the robot.}
    \label{fig:framework}
    \vspace{-1em}
\end{figure*}

\section{Background: Hamilton-Jacobi Reachability} \label{sec:background}
Our approach builds upon Hamilton-Jacobi (HJ) reachability analysis \cite{mitchell2005time, margellos2011hamilton}. 
This framework provides robust assurances, yields minimally invasive safety filters compatible with any nominal robot policy (e.g., a neural network),
nonlinear systems, and non-convex safety constraints.
Here we provide a brief background on the key components of HJ reachability and how to synthesize safety filters with this technique (see these surveys for more details \cite{bansal2017hamilton, wabersich2023data}). 

\para{Computing the Safety Filter} Given a failure set, $\failureset$, and the robot dynamics, HJ reachability computes a backward reachable tube (BRT), $\stateSpace^\dagger \subset \stateSpace$, which characterizes the set of initial states from which the robot is doomed to enter $\failureset$ despite its best control effort. 
The computation of the BRT can be formulated as a zero-sum, differential game between the control and disturbance, where the control attempts to avoid the failure region, whereas the disturbance attempts to steer the system inside it. 
This game can be solved using dynamic programming which, ultimately, amounts to solving the Hamilton Jacobi-Isaacs Variational Inequality (HJI-VI) \cite{fisac2015reach, margellos2011hamilton} to compute the value function $\valfunc$ that satisfies
\begin{equation}
\label{eq:HJIVI_BRS}
\begin{aligned}
\min \{&D_{\tdummy} \valfunc(\tdummy, \state) + H(\tdummy, \state, \nabla \valfunc(\tdummy, \state)), \marginfunc(\state) - \valfunc(\tdummy, \state) \} = 0 \\
&\valfunc(0, \state) = \marginfunc(\state), \quad \tdummy \le 0.
\end{aligned}
\end{equation}
Note that the function $\marginfunc(\state)$ is the implicit surface function representing our failure set $\failureset = \{\state: \marginfunc(\state) \le 0\}$.  
Here, $D_{\tdummy} \valfunc(\tdummy, \state)$ and $\nabla \valfunc(\tdummy, \state)$ denote the time and spatial derivatives of the value function $\valfunc(\tdummy,\state)$ respectively.
The Hamiltonian, $H(\tdummy, \state, \nabla \valfunc(\tdummy, \state))$, encodes the role of system dynamics, robot control, and disturbance, and is given by
\begin{equation}
\label{eq:HJIVI_ham}
H(\tdummy, \state, \nabla \valfunc(\tdummy, \state)) = \max_{\aR \in \ctrlSetR} \min_{\dstbR \in \dstbSetR} \nabla \valfunc(\tdummy, \state) \cdot \dyn(\state, \aR, \dstbR).
\end{equation}
The HJI-VI in \eqref{eq:HJIVI_BRS} can be solved offline via a variety of numerical tools, such as high-fidelity grid-based PDE solvers \cite{mitchell2004toolbox} or neural approximations that leverage self-supervised learning \cite{bansal2021deepreach} or adversarial reinforcement learning \cite{hsu2023isaacs}. 
Once the value function $\valfunc(\tdummy, \state)$ is computed, the BRT can be extracted from the value function's sub-zero level set
\begin{align}
\stateSpace^\dagger(\tdummy) = \{\state: \valfunc(\tdummy, \state) \le 0\}, 
\label{eq:unsafeSet} 
\end{align}
As $\tau \rightarrow -\infty$, the BRT represents the infinite time control-invariant set (denoted $\stateSpace^\dagger$ here on), which is what we use to construct the safety filter. Importantly, note that $\failureset \subseteq \stateSpace^\dagger$. 

\para{Shielding the Robot's Nominal Planner} 
Along with $\stateSpace^\dagger$, HJ reachability yields a corresponding policy-agnostic safety feedback controller $\fallback(\state)$ that guarantees to keep the robot \textit{outside} the BRT and \textit{inside} the safe states, $\safeSet = (\stateSpace^\dagger)^\mathsf{c}$. 
\begin{equation}
\label{eq:opt_ctrl}
\fallback(\state) =  \arg \max_{\aR \in \ctrlSetR} \min_{\dstbR \in \dstbSetR} \nabla \valfunc(\state) \cdot \dyn(\state, \aR, \dstbR),
\end{equation}
where $\valfunc(\state)$ represents the value function as 
$\tau \rightarrow -\infty$.
Using this, we can design a minimally-invasive control law (i.e., \textit{safety filter}) that \textit{shields} $\policyR$ from danger:
\begin{equation} \label{eqn:safe_control_law_0}
\overallpolicyR(\state)=
    \begin{cases}
      \policyR(\state; \env),& \text{if}~\state \in \safeSet\\
      \fallback(\state),& \text{otherwise}.
    \end{cases}
\end{equation}

\section{Updating Robot Safety Representations \\ Online from Natural Language Feedback} \label{sec:methodology}

While foundational safe control methods like the one in Section~\ref{sec:background} are powerful, they assume that the robot's safety representation (i.e., the failure set $\failureset$) is perfectly specified a priori. 
Our key idea is that vision-language models (VLMs) are not only a useful interface for people to specify unique constraints that they care about, but provide a flexible way to automatically convert multimodal data (vision and language) observed \textit{online} into constraint representations that are compatible with safe control tools. 
In this section, we detail the core components of our framework (in Figure~\ref{fig:framework}): (1) a VLM-based approach to updating the failure set and (2) an efficient warm-starting approach to update the safety filter online.

\para{Updating the Failure Set Online from Natural Language}
We design a constraint predictor 
\begin{equation}
    \pred(\obs^{0:t}, \langCmd^{0:t}, \hat{\failureset}^{t}_\env; \env) \rightarrow \hat{\failureset}^{t+1}_\env,
    \label{eq:constraint-predictor}
\end{equation}
that updates the inferred failure set based on the sequence of robot observations, human language commands, and last inferred failure set. 
We assume the initial inferred failure set, $\hat{\failureset}^0_\env$, is given to us; e.g., from mapping the robot's operating environment by running an off-the-shelf SLAM algorithm and extracting an occupancy map.
The core of our constraint predictor is a VLM that takes the current image observation ($\obs_t$) and the concatenation of all the human's language commands ($\langCmd_{0:t}$) so far, and produces bounding boxes ($\bbox_\image$) in the robot's image space associated with the language commands\footnote{Note that the bounding box is an over-approximation. Future work should use semantic segmentation for tighter failure constraint inference.}:
\begin{equation}
    \vlm(\obs^t, \langCmd^{0:t}) \rightarrow \bbox_\image.
\end{equation}
Utilizing the depth information from the RGB-D image $\obs_t$, these bounding boxes are projected into the ground plane via:
\begin{equation}
\text{proj}(\bbox_\image; \lambda) \rightarrow \bbox_{XY} \subset \stateSpace
\label{eq:pixel_to_world}
\end{equation}
where $\text{proj}(\cdot)$ is the standard camera projection operation depending on the camera intrinsics ($\lambda$) that we assume to be known. 
The predicted failure set, $\hat{\failureset}^{t+1}_\env$, is the prior failure set augmented with $\bbox_{XY}$. 
In total, $\pred$ is the composition of the VLM $\vlm$ and the operations for converting and augmenting the failure set: $\pred(\cdot, \cdot, \failureset^t_\env; \env) := \hat\failureset^t_\env \cup (\text{proj} \circ \phi)(\cdot, \cdot)$.

\para{Updating the Safety Filter Online via Warm Starting}
Every time the predicted failure set changes, $\hat{\failureset}^{t+1}_\env$, we need to also update the corresponding safety controller, ${\fallback}^{,t+1}$. 
However, this presents a computational challenge as we need to re-compute online a \textit{new} safety value function $\valfunc^{t+1}(\state)$ (in Equation~\ref{eq:HJIVI_BRS}) so that the robot always has a valid safety-preserving control law. 
To tackle this, we leverage the approach of warm-starting from \cite{bajcsy2019efficient, herbert2019reachability}.  
The intuition behind this approach is that since the failure set changes \textit{incrementally} and in a smaller region of the state space, the robot's corresponding safety value function should also only change in a smaller region of the state space. 
Prior work has precisely demonstrated this property, where warm-starting enables significantly faster updates of the BRT because fewer state values have to be updated \cite{bajcsy2019efficient}.
Thus, we leverage the value function computed at the prior timestep ($t$) to bootstrap the computation of the new value function ($t+1$)
by initializing $\valfunc^{t+1}(0, \state) = \valfunc^t(\state)$ (instead of the typical $\marginfunc(\state)$) in Equation~\ref{eq:HJIVI_BRS}. 

\section{Experimental Setup} \label{sec:setup}

\para{Robot Dynamics}
In simulation and hardware experiments, we model the robot as a 3D unicycle where the robot controls the linear and angular velocity:
\begin{equation} \label{eqn:3D_dubins_dyn}
\begin{aligned}
\dot{p}_x = v\cos\theta + \dstbR_x,\quad \dot{p}_y = v\sin\theta + \dstbR_y, \quad \dot{\theta} = \omega, 
\end{aligned}
\end{equation}
where $(p_x, p_y)$ is the planar position, $\theta$ is the heading, and $v$ is the speed of the robot.
The robot controls $\aR := (v, \omega)$ and for reachability analysis, we also model the disturbance to ensure a robust safety filter, $\dstbR = (\dstbR_x, \dstbR_y)$. 
In simulation, sim we used $\SI{0.1}{\meter\per\second} \leq v \leq \SI{1}{\meter\per\second}$, $|\omega| \leq \SI{1}{\radian\per\second}$ and $|\dstbR_i| \leq \SI{0.1}{\meter}, i \in \{x,y\}$.
In hardware, we changed the robot linear velocity bounds to $\SI{0.0}{\meter\per\second} \leq v \leq \SI{0.5}{\meter\per\second}$.

\begin{figure*}[t!]
    \centering
    \includegraphics[width=1\linewidth]{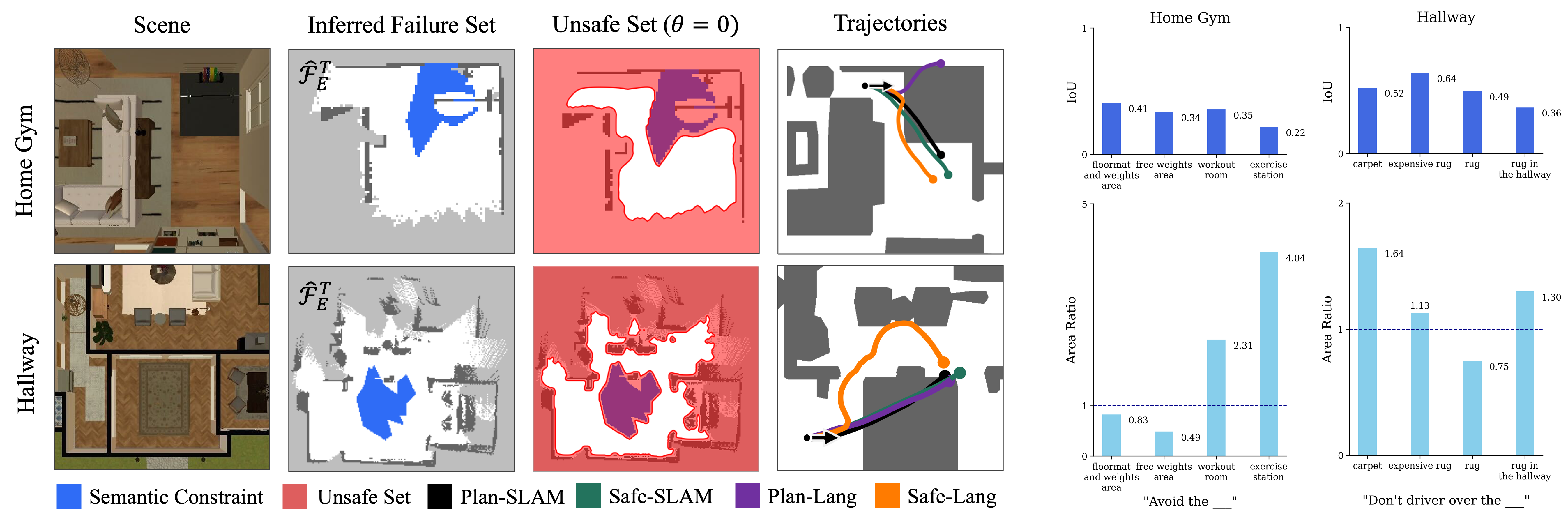}
    \caption{\textbf{Simulation: Closed-Loop Behavior.}  (left) Two simulated scenes from HSSD-HAB dataset \cite{minderer2022simple}, the final physical and semantic failure set and corresponding unsafe set, and the closed-loop trajectories of all methods. (right) Failure set inference accuracy as function of language command. Metrics compare the ground-truth failure $\failureset^*_\env$ set and the inferred failure $\hat\failureset^T_\env$.}
    \vspace{-1em}
    \label{fig:sim_results}
\end{figure*}

\para{Deployment Scenarios ($\env$)} 
We first deploy our framework the Habitat 3.0 simulator \cite{puig2023habitat3} in two different environments from the HSSD-HAB home dataset \cite{minderer2022simple}. 
The \textbf{home gym} scenario features a workout zone consisting of a floor mat and set of barbells in the corner of a living room (top left, Figure~\ref{fig:sim_results}).
The person wants the robot to avoid this area; e.g., because they are working out there or because they want the mat to stay clean. 
The \textbf{hallway} scenario features an expensive rug in the center of the room that the person doesn't want dirtied (bottom left, Figure~\ref{fig:sim_results})\footnote{We modified the original map slightly by removing the center bench.}. 
The rugs, workout mat, and weights pose a challenge for standard SLAM systems since their geometry alone is not sufficient to distinguish them from free-space. 
Instead, their subjective value to a person renders them a \emph{semantic} constraint that is communicated verbally.

\para{VLM Model ($\vlm$)}
In both simulated and hardware experiments, we use a pre-trained OWLv2 VLM \cite{minderer2024scalingopenvocabularyobjectdetection} which is an open-vocabulary object detector capable of identifying uncommon objects from natural language descriptions.

\para{Nominal Robot Policy ($\policyR$)}
Our approach is agnostic to the nominal robot policy. 
However, in experiments we use a Model Predictive Path Integral (MPPI) planner \cite{williams2016mppi}. 
The cost function consists of a goal-reaching term (sum of the Euclidean distance to a goal location, called \textit{cost to goal}) and a collision cost term (where, given the map of obstacles, the robot receives a high penalty for entering the obstacle zone and zero otherwise). 
Note that the MPPI planner does not model the disturbance in the dynamics; $\dstbR = 0$ in Equation~\ref{eqn:3D_dubins_dyn}.

\para{Methods} We compare two ways of inferring the failure set (SLAM-only vs. VLM-informed) and two robot policy  designs (with and without a safety filter). 
We use the RTAB-Map SLAM module \cite{labbe2019rtabmap}.
In total, we compare four methods: (1) \mppi: MPPI planner without a safety filter that plans around obstacles detected only by a SLAM module, (2) \mppilang: MPPI planner without a safety filter that plans around obstacles inferred by our VLM constraint inference predictor, (3) \safe: MPPI planner shielded by a safety filter that only knows of obstacles detected by SLAM, (4) \safelang: our approach that uses a language-informed safety filter.

\para{Deployment Details} 
We always keep the robot start and goal fixed.
When projecting the semantic constraint detections to the ground plane (Equation~\ref{eq:pixel_to_world}), we only include pixels within a distance threshold $\distThresh$ from the robot, ensuring that distant, free-space pixels are not incorrectly treated as part of the obstacle.
All modules run on individual threads, and the VLM and BRT run asynchronously on a NVIDIA RTX A6000.
To address delays in action execution or the network, we apply the safety filter at a small super-zero level set (i.e. a slight under-approximation of the safe \emph{zero} level set in Equation~\ref{eqn:safe_control_law_0}). 

\section{Simulation Results}
\label{sec:simulation-results}

\subsection{On the Accuracy of Failure Set Inference from Language}
\label{subsec:results-failure-inf}

For the same constraint,  users may give varying language descriptions.
Thus, we first study the accuracy of our VLM-based failure set inference to varying language inputs.

\para{Independent Variables} 
We test language commands with varying levels of detail about the failure set $\failureset^*_\env$. 
In the \textbf{home gym} scenario, the language follows a template: $\langCmd$ = ``\texttt{Avoid the} X'' where we vary X=\{\texttt{floormat and weights}, \texttt{free weights area}, \texttt{workout room}, \texttt{exercise station}\}. 
In the \textbf{hallway} scenario, the language follows a template: $\langCmd$ = ``\texttt{Don't drive over the} X'' where we vary X=\{\texttt{carpet}, \texttt{expensive rug}, \texttt{rug}, \texttt{rug in the hallway}\}. 

\para{Metrics} 
We compare the ground-truth $\failureset^*_\env$ obtained in the simulator to the final inferred $\hat{\failureset}^T_\env$ at the end of deployment. We measure the  \textbf{\textit{IoU (Intersection over Union)}} = $\frac{|\hat{\failureset}^T_\env \cap \failureset^*_\env|}{|\hat{\failureset}^T_\env \cup \failureset^*_\env|}$ which measures the accuracy of the inferred failure set by quantifying the alignment with the ground truth failure set. The closer to $IoU = 1$, the more accurate. 
We also measure \textbf{\textit{Area Ratio}} = $\frac{|\hat{\failureset}^T_\env|}{|\failureset^*_\env|}$ which measures how over-conservative (ratio $ > 1$) or under-conservative (ratio $ < 1$)  the inferred failure set is.

\para{Results} 
Figure \ref{fig:sim_results}  (right) shows the IoU and area ratio results for both the home gym and hallway scenarios. 
In the \textbf{home gym}, we find that as the language command becomes more vague, the VLM becomes \textit{over-conservative}, detecting a majority of the room as the constraint rather than just the floormat workout area (0.83 when commanded ``floormat and weights'' while 4.04 when commanded ``exercise station''). However, in the \textbf{hallway} scenario, the area ratio is fairly consistently close to 1.
Across both methods, however, the IoU scores are relatively low. 
This is because we only project pixels within the threshold distance, $\distThresh$, from the VLM detections onto $\hat{\failureset}_\env$. 
Thus, we tend to include less of the very distant parts of the failure set in our inferred set.
In practice, however, we found that this 
does not severely impact the robot’s 
behavior, which largely relies on reliable detection \emph{nearby}.

\begin{table*}[ht!]
    \centering
    \resizebox{\textwidth}{!}{%
    \begin{tabular}{l|c c c c|c c c c}
        \toprule
        & \multicolumn{4}{c|}{\textbf{Home Gym Scenario}} & \multicolumn{4}{c}{\textbf{Hallway Scenario}} \\
        Method & \multicolumn{1}{c}{Plan Time (s)} & \multicolumn{1}{c}{Cost to Goal} & \multicolumn{1}{c}{Abides $\failureset^*_\env$} & \multicolumn{1}{c|}{$\fallback$ Active } & \multicolumn{1}{c}{Plan Time (s)} & \multicolumn{1}{c}{Cost to Goal} & \multicolumn{1}{c}{Abides $\failureset^*_\env$} & \multicolumn{1}{c}{$\fallback$ Active } \\
        \midrule
        \mppi & 0.033 ($\pm$ 0.011) & 1.83  & \xmark & N/A & 0.083 ($\pm$0.053) & $3.39$ & \xmark & N/A \\
        \safe & 0.071 ($\pm$0.045) & 1.95 & \xmark & 0.0\% & 0.090 ($\pm$1.34) & 3.40 & \xmark & 12.37\% \\
        \mppilang & 0.115 ($\pm$0.054) & $\infty$ (collision) & \xmark & N/A & 0.056 ($\pm$0.055) & 3.19 & \xmark & N/A \\
        \safelang (ours) & 0.105 ($\pm$0.057) & 2.10 & \blackcheck & 23.83\% & 0.172 ($\pm$0.099) & 4.06 & \blackcheck & 61.67\% \\
        \bottomrule
    \end{tabular}
    }
    \vspace{0.1em}
    \caption{\textbf{Simulation: Closed-Loop Metrics.} Our approach consistently respects physical and semantic constraints. }
    \label{tab:sim-closed-loop-metric}
    \vspace{-1.em}
\end{table*}

\subsection{On the Closed-Loop Robot Performance} 
\label{subsec:results-closed-loop}

We next study the closed-loop performance of a robot navigating through our scenarios when using each method: $\mppi$, $\safe$, $\mppilang$, and $\safelang$. The language command is always kept the same (\textbf{home gym} is ``\texttt{Avoid the free weights area}'' and \textbf{hallway} is ``\texttt{Don't drive over the rug}'') and is given at $t = 0$.

\para{Metrics} 
We measure the speed of generating a robot action via the average \textbf{\textit{Plan Time}}. 
For \mppi and \mppilang it is the plan time required by MPPI, whereas for \safe and \safelang it includes the plan time of MPPI and the safety filtering. 
Note that the VLM calls and BRT updates are computed asynchronously, so they do not contribute to this metric.
We measure the goal reaching efficiency via  the average \textbf{\textit{Cost to Goal}} over the executed trajectory, where lower means more efficient. 
We also measure an indicator \textbf{\textit{Abides $\failureset^*_\env$}} if the robot ever violates $\failureset^*_\env$, 
and report \textbf{\textit{$\fallback$ Active}} for \safe and \safelang to measure the $\%$ of time the safety controller intervened during the trajectory.

\para{Results: Quantitative \& Qualitative} 
Table \ref{tab:sim-closed-loop-metric} shows quantitative results in both scenarios. 
Among the four methods, only ours was able to avoid both physical and semantic constraints. The planning time is not significantly increased by calling the safety filter, but the robot is slightly less efficient at goal reaching. 
We show qualitative results of closed-loop robot performance in Figure~\ref{fig:sim_results}. 
We observe that \mppi and \safe reach the goal while respecting the physical constraints, but completely violate the semantic constraints, as they can't be detected by SLAM alone. 
When language is included, \mppilang fails to adapt and ignores the semantic constraints detected in runtime: it either fails to find a feasible alternative path and ends up colliding (see \textbf{home gym} in Figure~\ref{fig:sim_results}), or slows down until it can't find an alternative path and move towards the goal ignoring the semantic constraint whatsoever (see \textbf{hallway} in Figure~\ref{fig:sim_results}). 
In contrast, our approach \safelang ensures that the robot will execute the optimal control action to avoid \textit{both} the semantic constraints and the physical constraints, so long as the BRT is updated fast enough. We study this more in Section~\ref{subsec:results-language-timing}.

\begin{figure}[h!]
    \centering
    \includegraphics[width=0.45\textwidth]{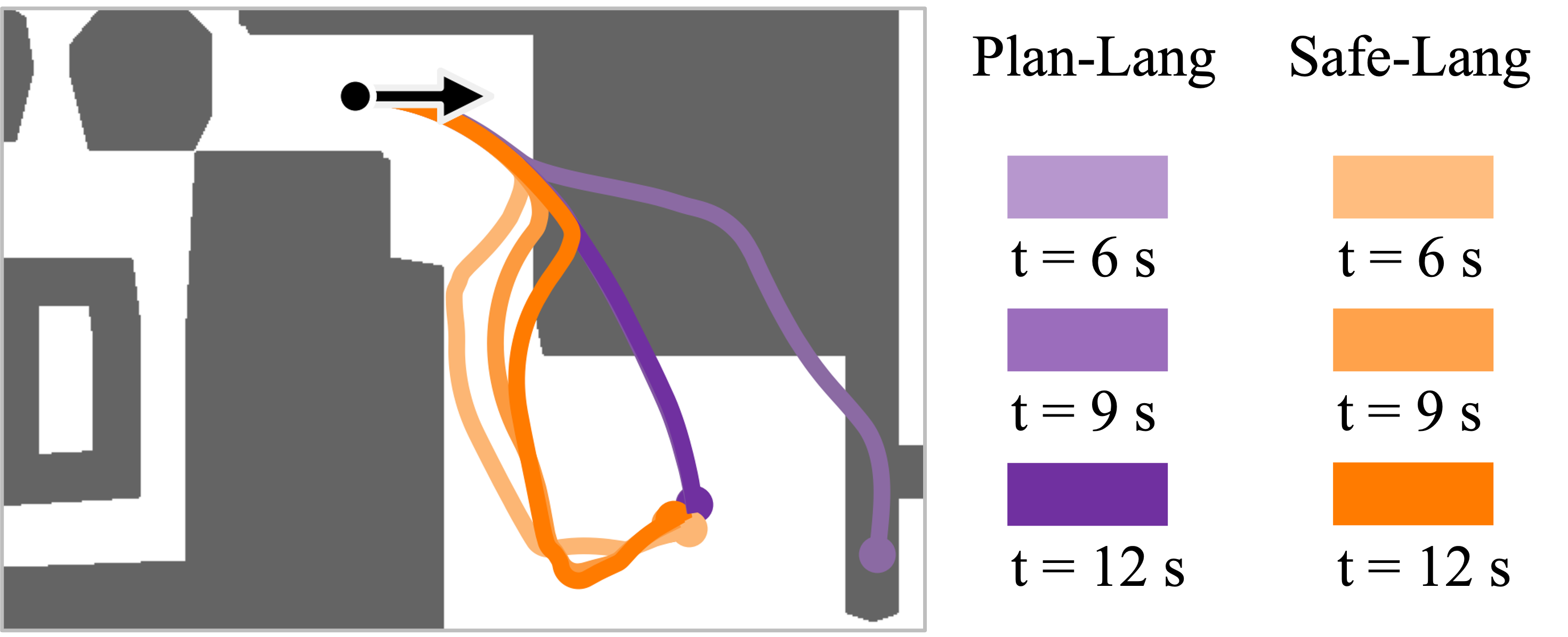}
    \caption{\textbf{Simulation: Language Timing.} Our \safelang method is more robust to feedback timing than \mppilang.}
    \label{fig:time-point-ablation}
    \vspace{-1.em}
\end{figure}

\subsection{On the Robustness to Language Feedback Timing}
\label{subsec:results-language-timing}

Next, we study the robustness of \safelang to language constraints added at some time $t>0$ during deployment. 
For brevity, we present results only in \textbf{home gym}. 

\para{Independent Variables}
We use the language command, $\langCmd^t$ = ``\texttt{Avoid the floor mat and weights}'', but vary the time when it is specified to the robot: $t = \{6s, 9s, 12s\}$.

\para{Results} 
Figure~\ref{fig:time-point-ablation} shows the closed-loop trajectories of the methods that use language feedback across all language command time points. 
Similar to prior studies \cite{trevisan2024biased}, we found that \mppilang fails to avoid constraints when the language feedback is obtained near the boundary of the constraint, especially when the free passage is narrow as in this study.
If the language constraint is added when the robot is closer to the rug ($t=9$), \mppilang fails to find a way around it (the robot slows down, but can't turn fast enough to avoid).
In contrast, our approach \safelang is more robust to language command timing. 
As long as the language command is given early enough so that the BRT can be updated in time (which took $~\SI{3}{\second}$ in this specific study), our method is able to avoid the new semantic constraints. 
Even when the robot was not able to completely avoid the constraint due to timing 
(as in $t = 12s$) our framework ensures the robot always has a best-effort action to leave the unsafe set as fast as possible.

\section{Hardware Experiments}
\label{sec:hardware-results}

\begin{figure}[t!]
    \centering
    \includegraphics[width=0.45\textwidth]{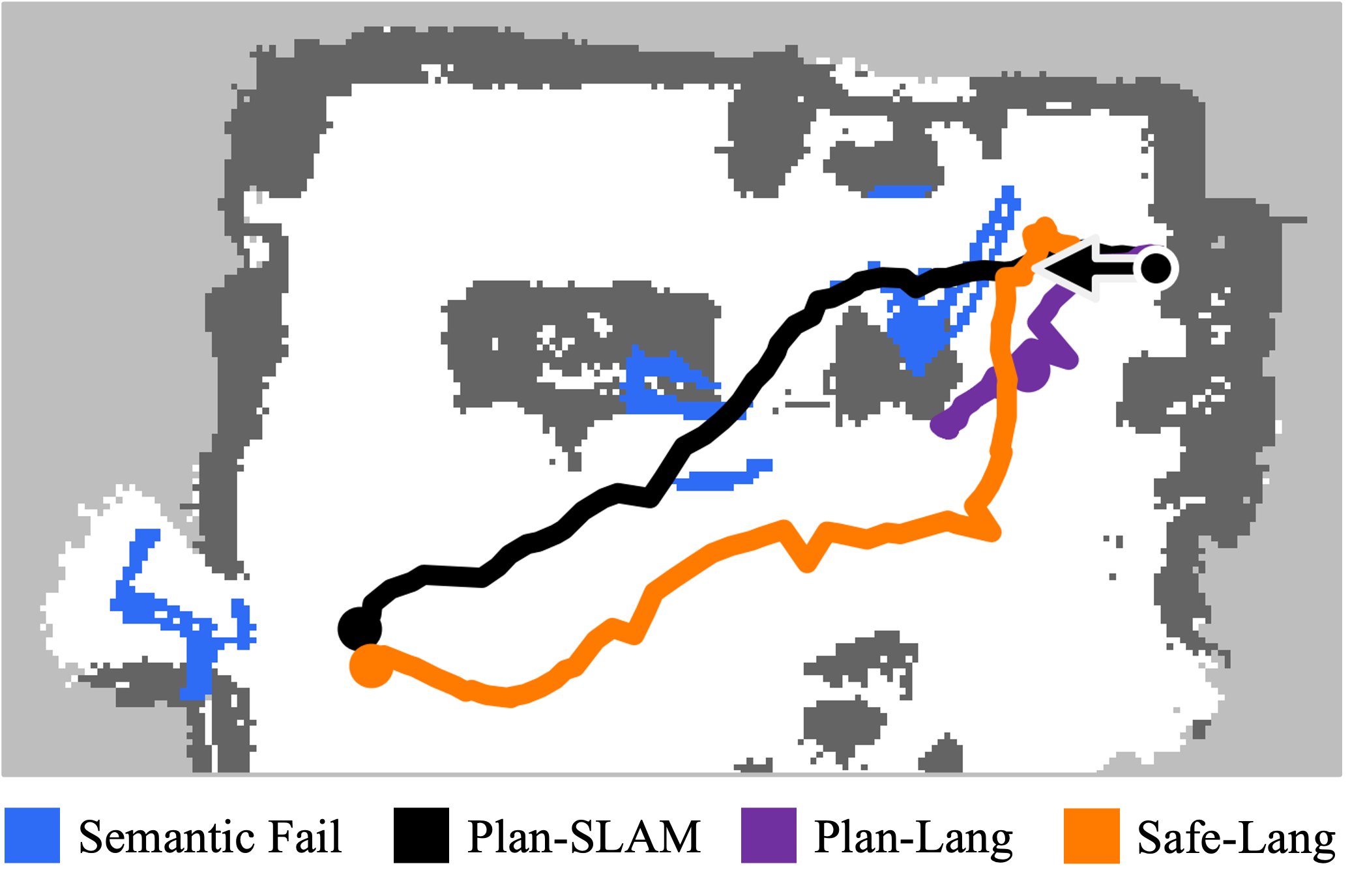}
    \caption{\textbf{Hardware: Closed-Loop Motion.} Without semantic constraints, \mppi cuts through the caution tape zone. \safelang respects both the physical and semantic constraints.}
    \label{fig:hardware-rollouts}
    \vspace{-1em}
\end{figure}

We deployed our framework in hardware on a LoCoBot ground robot equipped with an Intel RealSense camera. 

\para{Deployment Scenarios} 
We study a scenario that a real robot may face but is hard to simulate: avoiding areas marked by \textbf{caution tape}. The person specifies their desired constraint via the utterance $\langCmd^t$ = ``\texttt{Avoid the area surrounded by caution tape}'' at $t=0$ of robot deployment. We also qualitatively test a scenario with \textit{both} caution tape and coffee spill language constraints (Figure~\ref{fig:front-fig}) and another scene with $\langCmd^t=$``\texttt{Avoid the dog toys and the laundry}''. Videos are on the project website.

\para{Metrics} We use the same metrics as in Section~\ref{subsec:results-closed-loop} except we measure Time-to-Goal (in s) and Plan Time (in ms).

\begin{table}[t!]
    \centering
    \begin{tabular}{l|c c c c}
        \toprule
        & \multicolumn{4}{c}{\textbf{Caution Tape Scenario}} \\
        Method & \multicolumn{1}{c}{Plan Time (ms)} & \multicolumn{1}{c}{$t$-to-Goal } & \multicolumn{1}{c}{Abides $\failureset^*_\env$} & \multicolumn{1}{c}{$\fallback$ On } \\
        \midrule
        \mppi & 7 ($\pm$1) & 17.647  & \xmark & N/A \\
        \mppilang & 40 ($\pm$30) & $\infty$ & \xmark & N/A \\
        \safelang  & 31 ($\pm$29) & 26.176 & \blackcheck & 29.37\% \\
        \bottomrule
    \end{tabular}
    \vspace{0.1em}
    \caption{\textbf{Hardware: Closed-Loop Metrics.} We see similar trends in hardware as in simulation: informing a safety controller with language enables efficient task completion while respecting both physical and semantic constraints. }
    \vspace{-0.9em}
    \label{tab:hardware-closed-loop-metric}
\end{table}

\para{Results: Quantitative \& Qualitative}
We observed that our method was the only able to avoid both physical and semantic constraints (see Table~\ref{tab:hardware-closed-loop-metric} and Figure~\ref{fig:hardware-rollouts}). While the base planner \mppi avoids physical constraints and reach the goal, it could not avoid the semantic constraint as it is not detectable by the SLAM system alone. The language informed base planner \mppilang was in fact able to detect and plan around the semantic constraint, but ended up colliding with the physical obstacles, since it provides no guarantees in terms of safety and is not robust to new constraints added in runtime. Our method \safelang was able to react and avoid both physical and semantic constraints and reach the goal safely due to its strong safety guarantees.

\section{Conclusion} 
\label{sec:conclusion}

We propose a framework that enables robots to continuously update their safety representations online from natural language feedback. 
Our core idea is that pre-trained vision-language models can easily convert multimodal sensor observations to novel constraints
compatible with safety-oriented control tools such as Hamilton-Jacobi reachability. 
Across a suite of simulation and hardware experiments in ground navigation, we show that this is a promising first step towards enabling robots to continually refine their understanding of safety. Since we are interested in robot safety, future work should calibrate the output of the VLM (e.g., \cite{dixit2024perceive}) to provide assurances on constraint inference.

\bibliographystyle{IEEEtran}
\bibliography{bibliography}

\end{document}